\documentclass{article}

\usepackage[final]{neurips_2026}

\usepackage{natbib}
\setcitestyle{authoryear, round}

\usepackage[utf8]{inputenc}
\usepackage[T1]{fontenc}
\usepackage{times}
\usepackage{microtype}
\usepackage{nicefrac}

\usepackage{amsmath,amssymb,amsfonts,amsthm,mathtools,bbm}

\usepackage[table]{xcolor}
\definecolor{LightYellow}{rgb}{0.9954,0.98665,0.94745}
\definecolor{LightGreen} {rgb}{0.96 ,0.995 ,0.975 }
\definecolor{mydarkred}  {rgb}{0.4  ,0    ,0     }
\definecolor{mydarkgreen}{rgb}{0    ,0.6  ,0     }
\definecolor{SelfColor}  {rgb}{0.913,0.443,0.196}
\definecolor{UrlColor}   {rgb}{0.7098,0.009,0.0}
\definecolor{RefColor}   {rgb}{0.082 ,0.376,0.510}

\usepackage{graphicx}
\usepackage{subcaption}
\usepackage{booktabs,multirow,colortbl}
\usepackage{float,stfloats,wrapfig,svg,epsfig}

\usepackage{algorithm}
\usepackage{algpseudocode}

\usepackage{enumitem}
\usepackage[textsize=tiny]{todonotes}
\usepackage{amsthm}

\usepackage[colorlinks,
            linkcolor=UrlColor,
            urlcolor =UrlColor,
            citecolor=RefColor]{hyperref}
\usepackage[capitalize,noabbrev]{cleveref}
\makeatletter
\renewcommand{\@toptitlebar}{}
\renewcommand{\@bottomtitlebar}{\vskip 0.15in}
\makeatother

\title{%
\raggedright
{\color{RefColor}
Understanding Diversity Collapse in RLVR \\
via the Lens of Overtraining
}%
}

\author{%
\begin{minipage}{0.99\textwidth}
\raggedright
{\bfseries
Suqin Yuan\textsuperscript{1} \quad
Jinkun Chen\textsuperscript{2} \quad
Jiyang Zheng\textsuperscript{3} \quad
Muyang Li\textsuperscript{1} \quad \\
Lei Feng\textsuperscript{2}\footnotemark[1] \quad
Dadong Wang\textsuperscript{4} \quad
Tao Xiang\textsuperscript{5} \quad
Tongliang Liu\textsuperscript{1}\thanks{Corresponding authors.} \quad
Bo An\textsuperscript{6}
}\\[0.5em]
{\normalfont\mdseries
\textsuperscript{1} Sydney AI Centre, The University of Sydney \quad
\textsuperscript{2} Southeast University \quad
\textsuperscript{3} Microsoft\\
\textsuperscript{4} Data61, CSIRO \quad
\textsuperscript{5} Chongqing University \quad
\textsuperscript{6} Nanyang Technological University
}
\end{minipage}
}

\renewenvironment{abstract}{%
  \vskip 0.075in
  \noindent{\large\bfseries Abstract}\par
  \vspace{0.5ex}
  \noindent
  \begin{minipage}{\textwidth}
  \raggedright
}{%
  \end{minipage}
  \vskip 14ex
}

\begin{document}

\maketitle

\begin{abstract}
Reinforcement learning with verifiable rewards (RLVR) has become a key approach for enhancing the reasoning abilities of large language models. However, RLVR often suffers from \emph{diversity collapse}: Pass@$1$ improves while high-$k$ Pass@$k$ degrades, which is viewed as a narrowing of the model's reasoning boundary.
We formalize this diversity collapse through the lens of \emph{overtraining}: once a problem's contribution to the reference metric has effectively saturated, further updates no longer expand what the model can solve but still concentrate probability mass on the trajectories favored by on-policy sampling. Under a standard setup with few rollouts per problem, even a single observed success places a problem in a nearly saturated regime for high-$k$ Pass@$k$, so most updates in standard RLVR are overtraining from the boundary perspective.
This perspective also suggests a reading of whether RLVR can expand the model's reasoning abilities beyond the base model: since RLVR is structurally biased against high-$k$ Pass@$k$, its aggregate decline does not by itself mean that no new reasoning gains occurred. Interventionally, restricting updates to problems with zero observed success lifts Pass@$256$ above the base model on difficult benchmarks; observationally, a non-trivial fraction of initially unsolvable problems become solvable during standard RLVR training.
Building on these findings, we propose \emph{Bayesian Boundary Gating} (BBG), which redirects optimization away from overtraining by estimating each problem's marginal contribution to the reasoning boundary. Across multiple reasoning benchmarks, BBG improves average Pass@$k$ across a wide range of $k$.
\end{abstract}

\section{Introduction}
Reinforcement learning with verifiable rewards (RLVR) has emerged as a particularly effective paradigm for improving the reasoning ability of large language models (LLMs) on tasks with objectively checkable outcomes \citep{shao2024deepseekmath,lambert2025tulu,Guo_2025}. Despite these advances, RLVR is also known to suffer from \emph{diversity collapse}: while Pass@$1$ improves, high-$k$ Pass@$k$ frequently degrades as training progresses \citep{yue2025does,zhu2025the,li2026the}. Since high-$k$ Pass@$k$ measures whether the model can find a correct solution over many attempts, this degradation is often viewed as a narrowing of the model's reasoning boundary \citep{tuyls2025representation,yue2025does}.
One line of explanation attributes diversity collapse to the self-reinforcing nature of on-policy RLVR updates \citep{mahdavi2025beyond,walder2025passk,chen2025pass,thrampoulidis2025advantage}: trajectories that are already more likely to succeed are sampled and reinforced more often, creating a sharpening bias that concentrates probability mass on a shrinking subset \citep{jin2025revisiting,fan2026sharpening,li2026the}. This sharpening bias, however, is a property of every RLVR update; it does not by itself explain when an update still expands what the model can solve and when it merely reshapes probability mass among already reachable solutions.

\begin{figure}[t]
\centering
\begin{subfigure}[t]{0.52\textwidth}
    \centering
    \includegraphics[width=\textwidth]{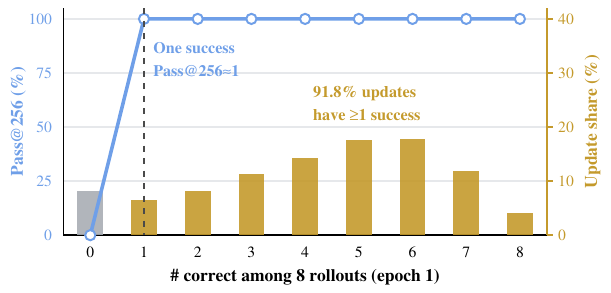}
    \vskip -0.06in
    \caption{Boundary saturation and update allocation}
    \label{fig:intro_boundary_overtraining}
\end{subfigure}
\hfill
\begin{subfigure}[t]{0.47\textwidth}
    \centering
    \includegraphics[width=\textwidth]{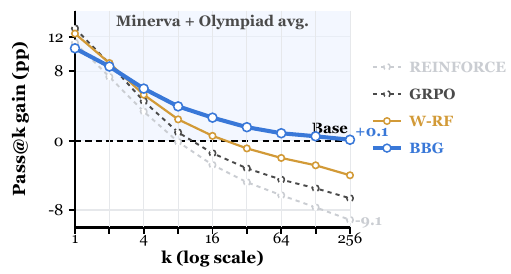}
    \vskip -0.06in
    \caption{Pass@$k$ gains over the base model}
    \label{fig:intro_bbg_passk}
\end{subfigure}
\vskip -0.01in
\caption{
(a) For $n=8$ rollouts, one observed correct rollout already places a problem in a nearly saturated regime for Pass@$256$, yet standard RLVR already allocates most updates even at epoch~1 to problems with at least one observed success.
(b) On two challenging benchmarks, BBG yields average positive Pass@$k$ gains across the evaluated range, whereas standard RLVR methods degrade at large $k$.
All methods are trained from Qwen2.5-Math-7B on the MATH training set, following the setup in Section~\ref{sec:late_stage}; (b) reports percentage gains over the base model using the results in Table~\ref{table:math_aime_pass_k}.
}
\label{fig:intro_overview}
\vskip -0.05in
\end{figure}

In this paper, we study this distinction through the lens of \emph{overtraining}: once a problem's contribution to a reference metric has effectively saturated, further updates can no longer expand what the model can solve. For high-$k$ Pass@$k$, this saturation is aggressive: since Pass@$k$ equals $1-(1-p)^k$ for a problem with success probability $p$, it saturates rapidly once $p$ becomes non-negligible. Under the standard few-rollout setup used in RLVR (e.g.\ $n=8$, \citealt{shao2024deepseekmath}), even a single correct rollout already yields an empirical success rate of $\hat{p}=1/8$, for which Pass@$256$ is $1-(7/8)^{256}=1-1.43\times10^{-15}\approx 1$. Thus, \emph{from the boundary perspective, most updates in standard RLVR are overtraining}: they can no longer meaningfully expand the reasoning boundary, but they can continue to sharpen the policy, leading to diversity collapse (Figure~\ref{fig:intro_boundary_overtraining}).

This perspective also offers a new reading of the widely reported aggregate decline of high-$k$ Pass@$k$ under RLVR, which has been viewed as evidence that RLVR does not expand reasoning capacity beyond the base model \citep{yue2025does}: if standard RLVR structurally overtrains on high-$k$ Pass@$k$, then an aggregate decline does not by itself mean that no new reasoning gains occurred. We provide further evidence that RLVR does expand the reasoning boundary during standard training, but this gain is masked by overtraining. Observationally, we track problems that are empirically unsolvable under the base policy (Pass@$256=0$) and find that a non-trivial fraction become solvable early in training, even as a larger number of previously solvable problems exit the boundary. Interventionally, restricting updates to problems with zero observed success, the only bucket where updates are not overtraining for high-$k$ Pass@$k$, lifts Pass@$256$ above the base model on difficult benchmarks, while higher-success buckets produce progressively larger declines.

Beyond the reasoning boundary, we also observe that current training recipes overtrain for Pass@$1$ itself: validation Pass@$1$ largely plateaus after the first few epochs, even as training-set success continues to climb throughout the remaining schedule. Together with the boundary-level overtraining identified above, this means that the later stages of standard RLVR training contribute little to Pass@$1$ while largely degrading high-$k$ Pass@$k$ (Figure~\ref{fig:passk_curves_boundary_avg}).
Building on this analysis, we propose \emph{Bayesian Boundary Gating} (BBG), a problem-level method that redirects optimization away from overtraining. BBG estimates each problem's expected marginal contribution to the reasoning boundary: problems with negligible boundary utility are excluded from updates entirely, while the remaining problems are weighted in proportion to their estimated contribution. Across multiple reasoning benchmarks, BBG improves average Pass@$k$ across a wide range of $k$ (Figure~\ref{fig:intro_bbg_passk}).

Our main contributions can be summarized as follows:
\begin{enumerate}[leftmargin=*,itemsep=1pt,topsep=2pt,parsep=2pt]
    \item We formalize diversity collapse in RLVR through the lens of \emph{overtraining} relative to a reference metric, and show that under standard rollout budgets, \emph{most updates are overtraining from the boundary perspective}.
    \item We provide both observational and interventional evidence that RLVR does expand the reasoning boundary, and that the widely reported aggregate decline of high-$k$ Pass@$k$ reflects overtraining on already-solved problems rather than an inability to acquire new reasoning capabilities.
    \item We propose \emph{Bayesian Boundary Gating} (BBG), which redirects optimization away from overtraining and improves average Pass@$k$ across a wide range of $k$ on multiple reasoning benchmarks.
\end{enumerate}
\newpage

\section{Related Work}

\textbf{RLVR.} Open systems such as Tulu 3, DeepSeekMath, DeepSeek-R1, and DAPO established the modern recipe space for RLVR \citep{lambert2025tulu,shao2024deepseekmath,Guo_2025,yu2025dapo}. A complementary line of work studies how to allocate limited training budget more effectively, including curriculum learning, selective rollouts, online difficulty filtering, and experience reuse \citep{chen2025self,zheng2025act,wang2025reinforcement,bae2026online,zhan2025exgrpo}. These works primarily target sample and training efficiency; by contrast, we take the evolving learning state of each problem and its consequence for high-$k$ Pass@$k$ as the central object of study.

\textbf{Diversity Collapse and the Reasoning Boundary.} Reinforcement learning from human feedback has long been observed to reduce output diversity \citep{kirk2024understanding,xiao2024algorithmic}. Within RLVR, this issue is especially consequential because reduced diversity directly lowers multi-attempt success. Recent work characterizes the phenomenon as reasoning boundary shrinkage or capability boundary collapse \citep{wen2025reinforcement,nguyen2025reasoning,dong2025rl,fan2026sharpening}, and \citet{yue2025does} uses the aggregate decline of high-$k$ Pass@$k$ to argue that RLVR does not expand reasoning capacity beyond the base model. Closely related, \citet{le2025no} and \citet{liu2025explore} revisit prompts with weak or zero within-group reward variance, showing that such prompts should not be uniformly discarded. We study diversity collapse from an overtraining perspective, linking it to both epoch-level and problem-level overtraining. We also argue that aggregate Pass@$k$ decline alone does not reliably indicate whether new reasoning capabilities have been acquired.

\textbf{Pass@$k$-aware Objectives.} Several recent methods revisit reward design, advantage estimation, or gradient reweighting to better preserve solution diversity and improve Pass@$k$ \citep{tang2025optimizing,mahdavi2025beyond,walder2025passk,chen2025pass,thrampoulidis2025advantage,peng2026simko}. \citet{yu2025pass} argues that Pass@$k$ is better viewed as a diagnostic of exploration than as a training objective. These works mostly design fixed surrogate objectives for Pass@$k$; we instead explore how the benefit of continuously optimizing Pass@$1$ for the reasoning boundary changes over training. We find that this benefit saturates quickly, while the sharpening of the policy continues.

\textbf{Over-optimization and Entropy Collapse.} Mismatches between proxy and true objectives can induce Goodhart-style degradation and reward hacking \citep{amodei2016concrete,skalse2022defining,gao2023scaling,karwowski2024goodharts}. In the RLVR setting, this concern manifests in entropy collapse, sampling bias, and divergence choice \citep{jin2025revisiting,fan2026sharpening,li2026the}. Existing mitigations regulate entropy or apply mass-covering regularization and alternative $f$-divergences to the high-success subset \citep{jin2025revisiting,nguyen2025reasoning,li2026the}; relatedly, emphasizing negative reinforcement can improve Pass@$k$ across a broad range of $k$ \citep{zhu2025the,yao2026the}. These approaches address diversity collapse through objective design or regularization. Our analysis further shows that overtraining for the reasoning boundary is present in most updates: under standard $n{=}8$ rollouts, even the smallest non-zero success already saturates Pass@$256$.

\subsection{A Simplified View of RLVR}
\label{sec:simplified_view}

Let $\mathcal{D}$ denote a distribution over training problems $x$, where each problem is instantiated as a prompt to the language model, and let $\pi_\theta$ be a language model policy over full response trajectories $y=(y_1,\ldots,y_T)$, where $\pi_\theta(y \mid x)$ denotes the probability of generating $y$ given prompt $x$. For tasks with verifiable outcomes, a deterministic verifier assigns a binary reward $r(x,y)\in\{-1,+1\}$, indicating whether the final response is correct. While practical RLVR implementations may additionally employ clipped surrogates, KL regularization \citep{schulman2017proximalpolicyoptimizationalgorithms}, or group-relative reward normalization \citep{shao2024deepseekmath}, the common signal shared across these methods is still induced by binary correctness under the current policy. We therefore isolate the verifier-induced success variable $s(x,y):=\frac{r(x,y)+1}{2}\in\{0,1\}$, and define the success probability of problem $x$ under the current policy as
\begin{equation}
\label{eq:prompt_success}
p_\theta(x):=\mathbb{E}_{y\sim\pi_\theta(\cdot\mid x)}[s(x,y)].
\end{equation}

Equivalently, if $\mathcal{C}(x):=\{y:\, s(x,y)=1\}$ denotes the set of correct trajectories for problem $x$, then
\begin{equation}
\label{eq:correct_set_mass}
p_\theta(x)=\sum_{y\in \mathcal{C}(x)}\pi_\theta(y\mid x).
\end{equation}
The verifier-induced component of RLVR can thus be written as
\begin{equation}
\label{eq:j_ver}
J_{\mathrm{ver}}(\theta)
:=\mathbb{E}_{x\sim\mathcal{D},\,y\sim\pi_\theta(\cdot\mid x)}[s(x,y)]
=\mathbb{E}_{x\sim\mathcal{D}}[p_\theta(x)].
\end{equation}
This is the average probability that the current policy solves a randomly drawn training problem. Denoting by Pass@$k$ the probability that at least one of $k$ independent samples is correct, $J_{\mathrm{ver}}(\theta)$ coincides with Pass@$1$. Because RLVR operates on-policy, each update tends to reinforce sampled correct trajectories and suppress incorrect ones, pushing probability mass onto the correct region and raising $p_\theta(x)$ in expectation \citep{schulman2017proximalpolicyoptimizationalgorithms,shao2024deepseekmath}.

\textbf{Rollout Success.}
During training, RLVR does not observe $p_\theta(x)$ directly. Instead, for each problem it only sees a finite number of on-policy rollouts. We therefore define the empirical success rate
\begin{equation}
\label{eq:empirical_success}
\hat p_{\theta,n}(x)
:=\frac{1}{n}\sum_{i=1}^{n}s(x,y_i),
\qquad y_i\sim \pi_\theta(\cdot\mid x),
\end{equation}
where $n$ is the number of sampled trajectories for problem $x$ in training. This statistic is quantized at resolution $1/n$. For example, when $n=8$, training can only distinguish the discrete success buckets $\{0/8,1/8,\ldots,8/8\}$.
For $m\in\{0,\ldots,n\}$, we define the corresponding rollout-success bucket
\begin{equation}
\label{eq:bucket_def}
B_{m/n}(\theta):=\left\{x:\hat p_{\theta,n}(x)=\frac{m}{n}\right\}.
\end{equation}
Since rollout success depends on the current policy, bucket membership changes as $\theta$ evolves during training, and we abbreviate $B_{m/n}(\theta)$ to $B_{m/n}$, e.g.\ $B_{0/8}, B_{1/8}, \ldots, B_{8/8}$ for $n=8$.


\section{Overtraining in RLVR}

We now connect \emph{diversity collapse} to \emph{overtraining} in RLVR. We begin by showing that validation Pass@$1$ saturates early in training, after which updates increasingly target near-resolved problems without producing corresponding gains in generalization (Section~\ref{sec:late_stage}). We then consider Pass@$k$ for large $k$ as a proxy for the model's reasoning boundary. Under a standard setup with $n=8$ rollouts, even the smallest non-zero observed success rate already saturates Pass@$k$, so most updates in standard RLVR are overtraining for the reasoning boundary (Section~\ref{sec:immediate}). Finally, we provide experimental evidence that RLVR can expand the reasoning boundary (Section~\ref{sec:reframe}).

\subsection{Overtraining for Pass@$1$}
\label{sec:late_stage}

Although optimizing $J_{\mathrm{ver}}$ effectively improves Pass@$1$, on-policy sampling combined with outcome-based advantage creates a self-reinforcing cycle: correct trajectories with higher probability under the current policy are sampled more often, receive reinforcement more frequently, and thereby further increase their share of the correct mass \citep{jin2025revisiting,fan2026sharpening,li2026the}. Over repeated updates, this concentrates $p_\theta(x)$ onto a shrinking subset of $\mathcal{C}(x)$, progressively sharpening the within-$\mathcal{C}(x)$ distribution even when the total correct mass $p_\theta(x)$ has long ceased to grow.

Following the practice of \citet{zhu2025the} (detailed in Appendix~\ref{app:training_details}), we train Qwen2.5-Math-7B \citep{yang2024qwen25mathtechnicalreportmathematical} on the MATH \citep{hendrycks2021measuring} training set for 20 epochs with standard RLVR (Eq.~\eqref{eq:j_ver}, $n=8$ rollouts per problem). As training progresses, the combined weight of the $B_{7/8}$ and $B_{8/8}$ buckets rises from $16.0\%$ to $64.3\%$, confirming that the majority of the training budget is eventually allocated to near-resolved problems where the correctness headroom $1-p_\theta(x)$ is small. This is consistent with recent work that addresses the high-success regime, whether by filtering prompts with near-perfect rollout success rates \citep{le2025no, liu2025explore} or by applying mass-covering regularization to the high-success subset \citep{li2026the, nguyen2025reasoning}.

\begin{figure}[t]
\centering
\vskip -0.05in
\begin{subfigure}[t]{0.33\textwidth}
    \centering
    \includegraphics[width=\textwidth]{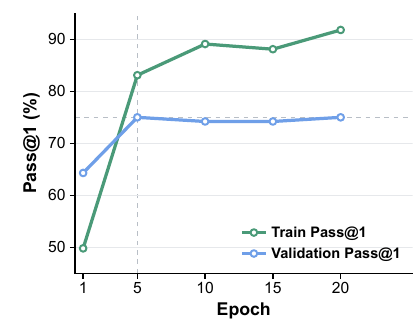}
    \vskip -0.1in
    \caption{ Pass@$1$ overtraining}
    \label{fig:pass1_overfitting}
\end{subfigure}
\hfill
\begin{subfigure}[t]{0.325\textwidth}
    \centering
    \includegraphics[width=\textwidth]{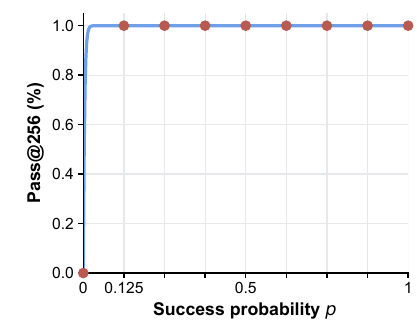}
    \vskip -0.1in
    \caption{ Pass@$256$ overtraining}
    \label{fig:boundary_saturation_curve}
\end{subfigure}
\hfill
\begin{subfigure}[t]{0.33\textwidth}
    \centering
    \includegraphics[width=\textwidth]{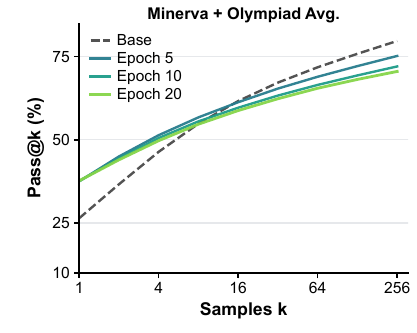}
    \vskip -0.1in
    \caption{ Large-$k$ degradation}
    \label{fig:passk_curves_boundary_avg}
\end{subfigure}
\vskip -0.05in
\caption{
Overtraining in RLVR. 
(a) Training Pass@$1$ continues to increase after validation Pass@$1$ has largely saturated.
(b) For $k=256$ and $n=8$, even the smallest non-zero empirical success bucket already saturates Pass@$256$.
(c) On MinervaMath \citep{lewkowycz2022solvingquantitativereasoningproblems} and OlympiadBench \citep{he2024olympiadbenchchallengingbenchmarkpromoting}, later epochs provide little additional Pass@$1$ gain but largely reduce large-$k$ Pass@$k$.
}
\label{fig:late_stage_overtraining}
\vskip -0.15in
\end{figure}

Further, we track training-set and validation performance during training. In early training (epochs 1--5), the training-set rollout success rate and validation Pass@$1$ (MATH-500; \citet{lightman2024lets}) both improve together. After epoch~5, however, a characteristic overfitting pattern emerges (Figure~\ref{fig:pass1_overfitting}): the training-set success rate climbs from $68.69\%$ to $77.06\%$, while validation Pass@$1$ plateaus at around $72\%$. This divergence suggests that the policy learns to solve the same training problems more reliably, but these updates no longer contribute to validation Pass@$1$ and instead only narrow the distribution over correct trajectories.
These observations suggest that overtraining for Pass@$1$ is not limited to diminishing returns on near-resolved buckets: overtraining also operates at the \emph{epoch level}. Under the planned training schedule of current recipes \citep{zhu2025the}, training continues well past the point where validation Pass@$1$ saturates, yet the policy continues to be optimized throughout. We argue below, however, that if we shift the reference objective from Pass@$1$ to Pass@$k$ for large $k$, the scope of overtraining turns out to be far broader.

\subsection{Overtraining for Pass@$k$}
\label{sec:immediate}

The analysis above takes $J_{\mathrm{ver}}=J_1$ (Pass@$1$) as the reference and identifies overtraining primarily in the near-resolved buckets and the late training epochs. We now show that if we regard Pass@$k$ for large $k$ as a proxy for the model's \emph{reasoning boundary} \citep{yue2025does}, the frontier of problems it can solve given sufficiently many attempts, then the scope of overtraining extends far beyond the near-resolved regime. From the reasoning-boundary perspective, most updates applied to a problem with $\hat p_{\theta,n}(x)\ge 1/n$, whether reinforcing correct rollouts or suppressing incorrect ones, is optimizing a problem that is already solved.
Define the Pass@$k$ objective over the training distribution as
\begin{equation}
\label{eq:j_k}
J_k(\theta)\;:=\;\mathbb{E}_{x\sim\mathcal{D}}\!\Big[1-\bigl(1-p_\theta(x)\bigr)^k\Big].
\end{equation}
This is the expected probability that at least one of $k$ independent samples solves problem $x$. The marginal value of a unit increase in $p_\theta(x)$ under this objective is
\begin{equation}
\label{eq:marginal_k}
\frac{\partial}{\partial\,p_\theta(x)}\Big[1-\bigl(1-p_\theta(x)\bigr)^k\Big] \;=\; k\,\bigl(1-p_\theta(x)\bigr)^{k-1},
\end{equation}
which decays exponentially once $p_\theta(x)$ exceeds a small threshold. In practice, RLVR typically uses a small number of rollouts per problem, $n=8$ \citep{shao2024deepseekmath} or $n=16$ \citep{Guo_2025}, so the training procedure can only observe success rates at a coarse resolution of $1/n$. Figure~\ref{fig:boundary_saturation_curve} makes the consequence concrete for $k=256$ and $n=8$. Recall from Eq.~\eqref{eq:empirical_success} that $\hat p_{\theta,n}(x)=m/8$ is not the true success probability $p_\theta(x)$ but a coarse empirical estimate based on only 8 rollouts. Nonetheless, using the plug-in estimate $p_\theta(x)=1/8$ for a problem that produced one correct rollout, the resulting $\text{Pass@}256(x)=1-(7/8)^{256}=1-1.43\times10^{-15}\approx 1$ is already effectively saturated. All subsequent buckets are equally saturated. The true $p_\theta(x)$ for a problem observed at $\hat p_{\theta,8}(x)=1/8$ may differ from exactly $1/8$, but for any $p_\theta(x)$ that is non-negligible enough to produce at least one success in 8 trials, the Pass@$256$ objective is already indistinguishable from~$1$.

This has a direct consequence for understanding why RLVR degrades high-$k$ Pass@$k$. From the reasoning-boundary perspective, standard RLVR provides essentially \emph{no positive contribution} to $J_k$ for large $k$: since $J_k$ is already saturated for every problem with $\hat p_{\theta,n}(x)\ge 1/n$, any update on these problems, regardless of the sign of the advantage, yields no boundary expansion, while still contributing to policy sharpening. The only bucket where an update could meaningfully improve $J_k$ is $B_{0/n}(\theta)$: problems that have not yet produced a single correct rollout (detailed in Appendix~\ref{app:zero_bucket}).

The two forms of \emph{overtraining} identified in Sections~\ref{sec:late_stage} and \ref{sec:immediate} appear together in practice: as shown in Figure~\ref{fig:passk_curves_boundary_avg}, on the MinervaMath+OlympiadBench average, training from epoch 5 to epoch 20 improves Pass@$1$ only from $37.3\%$ to $37.7\%$, while Pass@$256$ drops from $75.2\%$ to $70.6\%$. The per-benchmark Pass@$k$ curves are provided in Appendix~\ref{app:boundary_dynamics}. This leads to two key observations:
\begin{enumerate}[leftmargin=*,itemsep=1pt,topsep=2pt,parsep=2pt]
    \item \textbf{Current training schedules overtrain.} Validation Pass@$1$ saturates early in training, yet the policy continues to be optimized throughout the remaining epochs.
    \item \textbf{Most updates are overtraining for the reasoning boundary.} Updates on problems that have produced at least one correct rollout ($\hat p \ge 1/n$) cannot expand the reasoning boundary, regardless of whether they reinforce correct trajectories or suppress incorrect ones.
\end{enumerate}

\subsection{Does RLVR Expand the Reasoning Boundary?}
\label{sec:reframe}

The analysis above establishes that standard RLVR provides near-zero marginal benefit to $J_k$ for large $k$ on any problem that has already produced at least one correct rollout, while simultaneously harming it through \emph{overtraining}. A natural corollary is that overall high-$k$ Pass@$k$ is expected to decline under RLVR training, as a direct consequence of \emph{overtraining}.
This calls into question the framing adopted by recent work that uses the decline of high-$k$ Pass@$k$ (e.g.\ Pass@$256$) as evidence that RLVR does not incentivize reasoning capacity beyond the base model \citep{yue2025does}. If RLVR inherently suppresses this metric through \emph{overtraining} regardless of whether new capabilities are acquired, then \emph{aggregate} Pass@$256$ degradation does not indicate the absence of reasoning gain.

\begin{figure*}[h]
\vskip -0.2in
\centering
\begin{subfigure}[b]{0.51\textwidth}
    \centering
    \includegraphics[width=\textwidth]{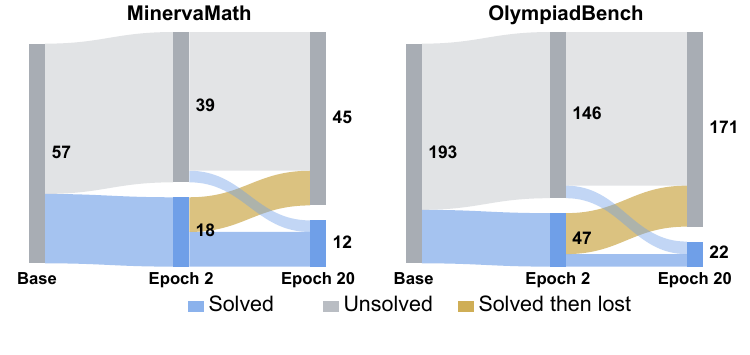}
    \vskip -0.16in
    \caption{Boundary across epochs}
    \label{fig:boundary_tracking_sankey}
\end{subfigure}
\hfill
\begin{subfigure}[b]{0.46\textwidth}
    \centering
    \includegraphics[width=\textwidth]{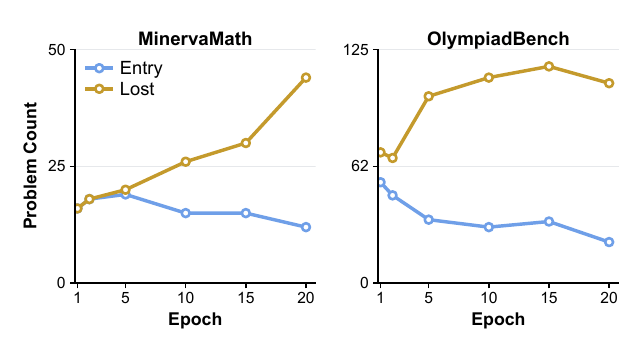}
    \vskip -0.15in
    \caption{Entry and loss across epochs}
    \label{fig:boundary_tracking_entry_loss}
\end{subfigure}
\vskip -0.07in
\caption{\emph{RLVR expands the reasoning boundary under standard training.}
(a) Sankey diagram tracking problems by Pass@$256$ status across epochs: a non-trivial fraction of initially unsolvable problems enter the boundary, but this gain is increasingly offset by problems exiting it. (b) Per-epoch entry and loss counts; entry peaks early while loss accumulates, driving the aggregate Pass@$256$ decline. See Appendix~\ref{app:sampling_noise} for a sampling-noise analysis. Additional results in Appendix~\ref{app:boundary_dynamics}.}
\label{fig:boundary_tracking}
\vskip -0.07in
\end{figure*}

\begin{wrapfigure}{r}{0.38\textwidth}
\vskip -0.42in
\centering
\includegraphics[width=\linewidth]{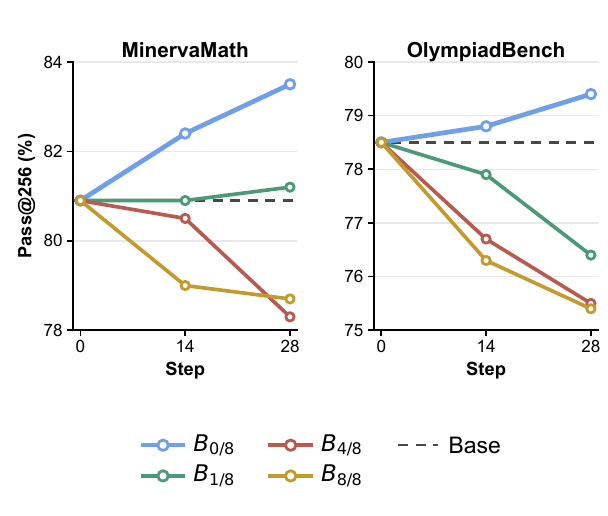}
\vskip -0.12in
\caption{Pass@$256$ improves when updates are restricted to $B_{0/8}$, and declines under higher-success buckets or full-data RLVR on harder benchmarks.}
\label{fig:bucket_restricted}
\vskip -0.3in
\end{wrapfigure}

We investigate this question with the same Qwen2.5-Math-7B setup used in Sections~\ref{sec:late_stage} and \ref{sec:immediate}, through two complementary lenses (Figures~\ref{fig:boundary_tracking} and \ref{fig:bucket_restricted}): \emph{observationally}, by tracking whether initially unsolvable problems become solvable under standard RLVR training; and \emph{interventionally}, by restricting updates to the $B_{0/8}$ bucket to isolate the boundary-expanding signal from overtraining.

\textbf{Tracking initially unsolvable problems.} For each benchmark, we identify the subset of problems with Pass@$256=0$ under the base policy (i.e.\ problems outside the initial reasoning boundary) and re-evaluate Pass@$256$ on this fixed subset at each training epoch of standard RLVR. As shown in Figure~\ref{fig:boundary_tracking}, a non-trivial fraction of these initially unsolvable problems become solvable during training on both an in-distribution math validation set and a held-out benchmark, demonstrating that RLVR does acquire new reasoning capacity on problems the base model cannot reach at all. This gain is not visible in aggregate Pass@$256$ because it is offset by a larger number of problems exiting the boundary as training progresses.

\textbf{Bucket-restricted RLVR.} To isolate which updates contribute to boundary expansion, we train variants in which updates are restricted to a single rollout-success bucket $B_{m/8}$ for $m\in\{0,1,4,8\}$, and compare Pass@$256$ to the base model. By the analysis of Section~\ref{sec:immediate}, only updates in $B_{0/8}$ carry non-negligible marginal value for $J_k$; updates in higher-success buckets can only sharpen the policy. As shown in Figure~\ref{fig:bucket_restricted}, restricting updates to $B_{0/8}$ improves Pass@$256$ over the base model on both benchmarks, while higher-success buckets can produce progressively larger declines. This is consistent with the analysis in Section~\ref{sec:immediate}: the source of aggregate Pass@$256$ degradation is not $J_1$ optimization itself, but the application of updates to problems whose $J_k$ contribution has already saturated.

Together, the two experiments confirm that RLVR does expand the reasoning boundary: some initially unsolvable problems become solvable, and isolating updates to $B_{0/8}$ raises Pass@$256$ above the base model on difficult benchmarks. The aggregate decline reported in prior work reflects a net outcome in which these gains are offset by losses on already-solvable problems. The analysis in Sections~\ref{sec:late_stage} and~\ref{sec:immediate} explains why this offset dominates: under standard $n{=}8$ rollouts, the vast majority of updates target problems whose $J_k$ contribution is already saturated, contributing only to policy sharpening. 

\section{Methodology}
\label{sec:method}

The above analysis suggests that \emph{diversity collapse} is better understood as a misallocation of optimization across learning states: standard RLVR applies the same update pressure to every problem regardless of whether it is already solved, and the resulting \emph{overtraining} drives the observed degradation in high-$k$ Pass@$k$. A natural and simple intervention is to reallocate optimization toward problems that are not yet \emph{resolved}. The difficulty is that the true success probability $p_\theta(x)$ is not observed; training only reveals the empirical success rate $\hat p_{\theta,n}(x)=m/n$ (Eq.~\eqref{eq:empirical_success}).

\textbf{Bayesian boundary utility.}
To assess how much an update on a given problem can still contribute to the reasoning boundary, we recall from Eq.~\eqref{eq:marginal_k} that the marginal value of raising $p_\theta(x)$ for $J_k$ is $k(1-p)^{k-1}$, but the true $p_\theta(x)$ is unobserved. We adopt a standard Beta-Binomial Bayesian model to handle this uncertainty: placing a uniform prior on $p$ and conditioning on the observed $m$ successes out of $n$ rollouts yields the posterior $p \mid m,n \;\sim\; \mathrm{Beta}(m{+}1,\, n{-}m{+}1)$. The expected marginal contribution of problem $x$ to $J_k$ under this posterior is
\vskip -0.2in
\begin{equation}
\label{eq:boundary_utility}
u(m;\,n,k)
\;:=\;\mathbb{E}_{p\,\sim\,\mathrm{Beta}(m+1,\,n-m+1)}\!\Big[k\,\bigl(1-p\bigr)^{k-1}\Big]
\;=\;k\,\prod_{j=1}^{k-1}\frac{n-m+j}{n+1+j}\,,
\end{equation}
\vskip -0.1in
where the closed form follows from the moments of the Beta distribution \citep{gelman1995bayesian}. Intuitively, $u(m;\,n,k)$ measures the expected marginal improvement in Pass@$k$ from a unit increase in $p_\theta(x)$, averaged over posterior uncertainty about $p$. For $n{=}8$ and $k{=}256$, the per-bucket utility drops sharply: $u(0;\,8,256)\approx 8.73$, while $u(1;\,8,256)\approx 0.27$ and $u(m;\,8,256)<10^{-2}$ for all $m\ge 2$. That is, $96.97\%$ of $\sum_{m=0}^{n}u(m;\,n,k)$ is concentrated on a single bucket $B_{0/8}$.

\textbf{Bayesian Boundary Gating (BBG).}
We use $u(m;\,n,k)$ to identify which problems are effectively resolved from the boundary perspective and suppress updates on them. Since $\frac{1}{n+1}\sum_{m=0}^{n}u(m;,n,k)=1$, the utility is naturally normalized. We set
\vskip -0.16in
\begin{equation}
\label{eq:bbg_weight}
\alpha(m) \;=\;
\begin{cases}
u(m;\,n,k), & \text{if } u(m;\,n,k) \;\ge\; \tau\,(n{+}1), \\[2pt]
0, & \text{otherwise},
\end{cases}
\end{equation}
\vskip -0.06in
where $\tau$ is a small threshold (we use $\tau{=}0.01$). Buckets whose share of the total utility falls below $\tau$ are hard-gated and excluded from backpropagation entirely. Because $\alpha$ is fixed given $(m,n,k)$ and does not depend on the batch composition, BBG's gating behavior is consistent regardless of the bucket distribution within a particular batch.

BBG is applied on top of the signed verifier loss (Sec.~\ref{sec:simplified_view}; $r\in\{-1,+1\}$) and only modulates the problem-level coefficient $\alpha(m)$ without altering the trajectory-level loss. In particular, for problems in $B_{0/n}$ all rollouts receive $r{=}{-1}$, so the update signal is purely negative reinforcement that suppresses sampled incorrect trajectories; this requires the advantage computation to assign nonzero values to uniform-outcome groups. Since BBG controls which problems are updated rather than how each problem is optimized, it is in principle complementary to methods that operate at the objective or trajectory level, such as Pass@$k$-aware gradient reweighting \citep{walder2025passk}, negative-reinforcement emphasis \citep{zhu2025the}, and pairwise ranking losses \citep{peng2026simko}.

\section{Experiments}
\label{sec:experiments}

\textbf{Settings.}
Our primary experiments use Qwen2.5-Math-7B \citep{yang2024qwen25mathtechnicalreportmathematical} as the base model, trained on the MATH training set \citep{hendrycks2021measuring} ($\approx$7.5k problems). Following the practice of \citet{zhu2025the}, we train with the verl framework \citep{10.1145/3689031.3696075} using $n{=}8$ rollouts per problem, a peak learning rate of $1 \times 10^{-6}$ with cosine decay, a training batch size of 1024, and mini-batch size of 128. The maximum prompt length is 1024 tokens and the maximum response length is 3072 tokens. Rollouts are generated with vLLM \citep{10.1145/3600006.3613165} in bfloat16 precision, and the KL coefficient is set to 0. All methods are trained for 140 steps ($\approx$20 epochs) on a homogeneous cluster equipped with 4 H100 GPUs per node, using FSDP and gradient checkpointing. For BBG, we use the Bayesian posterior estimator with $k{=}16$ as the boundary reference, which sits at the geometric midpoint of the evaluated range $k \in [1, 256]$, and set the minimum weight ratio $\tau{=}0.01$ (Eq.~\eqref{eq:bbg_weight}); the bucket mask accumulates problems into an effective batch of 512 active prompts per step. BBG is combined with W-RF \citep{zhu2025the}, where the weight of positive reward is reduced to $0.3$. Further details are provided in Appendix~\ref{app:training_details}.

\textbf{Training and evaluation.}
We compare against three baselines: (i)~\emph{REINFORCE}, which applies the standard policy gradient with binary reward $r \in \{-1, +1\}$; (ii)~\emph{GRPO} \citep{shao2024deepseekmath}, which normalizes advantages within each problem's rollout group; and (iii)~\emph{W-RF} \citep{zhu2025the}, which emphasizes negative reinforcement by reweighting the advantage to mitigate diversity collapse. For each evaluation benchmark, we generate 256 independent samples per problem using temperature $T{=}0.6$ and nucleus sampling with top-$p{=}0.95$, with a maximum generation length of 4096 tokens. Pass@$k$ is computed using the unbiased estimator of \citet{chen2021evaluatinglargelanguagemodels}. 

\textbf{Main Results.}
Table~\ref{table:math_aime_pass_k} presents Pass@$k$ results across four benchmarks spanning a range of difficulty. BBG attains the highest average Pass@$k$ on every benchmark, and is the only method whose Pass@$256$ matches or exceeds the base model on all four. Following the analysis in Section~\ref{sec:immediate}, we organize the discussion by benchmark difficulty: diversity collapse is likely to appear on benchmarks with many near-boundary problems, reachable under many samples but not yet reliably solved.

\begin{description}[leftmargin=0em, labelsep=0.5em, itemsep=3pt, topsep=-1pt, parsep=0pt]

\item[{\normalfont\itshape MATH-500}] \citep{lightman2024lets}. All methods improve Pass@$1$, and the high Pass@$256$ ceiling (base $96.2\%$) leaves little room for boundary shrinkage. The high-$k$ cost of overtraining is thus only faintly visible: REINFORCE, GRPO, and W-RF dip slightly below the base model at $k{=}256$, while BBG is the only method that improves Pass@$k$ at \emph{every} $k$. On in-distribution MATH-500, the cost of overtraining is largely hidden by saturation and ceiling effects.

\item[{\normalfont\itshape OlympiadBench \& MinervaMath}]
\citep{he2024olympiadbenchchallengingbenchmarkpromoting,lewkowycz2022solvingquantitativereasoningproblems}. Diversity collapse emerges most clearly on these intermediate-difficulty benchmarks. All methods improve Pass@$1$, but REINFORCE, GRPO, and W-RF all fall below the base model at high $k$, reflecting boundary shrinkage. W-RF \citep{zhu2025the} reduces this high-$k$ degradation through negative reinforcement but does not eliminate it. In contrast, BBG is the only method whose Pass@$256$ matches or exceeds the base model on both benchmarks, and it also attains the largest average Pass@$k$.

\item[{\normalfont\itshape AIME 2025}] \citep{zhu2025the}. W-RF and BBG improve Pass@$k$ across the whole evaluated range, whereas REINFORCE and GRPO still fall below the base model at high $k$. This does not contradict the overtraining account: AIME is sufficiently difficult that the $1$--$256$ range still probes a relatively low-success regime, so as $k$ increases Pass@$k$ saturates at progressively smaller $p_\theta(x)$, extending the overtraining regime to harder problems. This is consistent with \citet{yue2025does}, who report Pass@$k$ degradation on AIME at $k{=}1024$.

\end{description}

\newcommand{\up}[1]{{\scriptsize\textcolor{mydarkgreen}{\,\textsuperscript{+#1}}}}
\newcommand{\dn}[1]{{\scriptsize\textcolor{mydarkred}{\,\textsuperscript{--#1}}}}
\newcommand{\zer}{{\scriptsize\textcolor{gray}{\,\textsuperscript{$\pm$0}}}}

\begin{table*}[t]
\vskip -0.3in
\newcommand{\tabcite}[1]{{\tiny\citep{#1}}}
\centering
\setlength{\tabcolsep}{1.1pt}
\renewcommand{\arraystretch}{0.9}
\small
\caption{Pass@$k$ results with Qwen2.5-Math-7B under exact any-boxed answer scoring. \emph{Avg.}\ denotes the arithmetic mean over $k \in \{1,2,4,8,16,32,64,128,256\}$. \emph{Superscripts} indicate the change relative to the base model: \textcolor{mydarkgreen}{green} for improvement, \textcolor{mydarkred}{red} for degradation.}
\vskip -0.07in
\label{table:math_aime_pass_k}
\begin{tabular}{l*{9}{c}|>{\columncolor{RefColor!10}}c}
\toprule
\textbf{Method} & \multicolumn{9}{c}{\textbf{Pass@$k$}} & \cellcolor{white}\textbf{Avg.} \\
\cmidrule(lr){2-10}
$k$ & 1 & 2 & 4 & 8 & 16 & 32 & 64 & 128 & 256 & \cellcolor{white} \\
\midrule
\multicolumn{11}{c}{\textbf{MATH-500 \citep{lightman2024lets}}} \\
\midrule
Base Model & 64.5 & 76.2 & 83.8 & 88.2 & 90.8 & 92.6 & 94.0 & 95.3 & 96.2 & 86.8 \\
REINFORCE & 75.5\up{11.0} & 79.7\up{3.5} & 82.7\dn{1.1} & 85.9\dn{2.3} & 87.9\dn{2.9} & 88.8\dn{3.8} & 89.5\dn{4.5} & 91.0\dn{4.3} & 92.0\dn{4.2} & 85.9\dn{0.9} \\
GRPO \citep{shao2024deepseekmath} & \textbf{77.5}\up{13.0} & \textbf{81.7}\up{5.5} & 84.6\up{0.8} & 86.6\dn{1.6} & 88.3\dn{2.5} & 89.7\dn{2.9} & 91.0\dn{3.0} & 92.0\dn{3.3} & 93.0\dn{3.2} & 87.2\up{0.4} \\
W-RF \citep{zhu2025the} & 75.5\up{11.0} & 80.8\up{4.6} & 84.5\up{0.7} & 87.0\dn{1.2} & 88.9\dn{1.9} & 90.7\dn{1.9} & 92.2\dn{1.8} & 93.2\dn{2.1} & 93.8\dn{2.4} & 87.4\up{0.6} \\
BBG (Ours) & 75.6\up{11.1} & 81.6\up{5.4} & \textbf{86.3}\up{2.5} & \textbf{89.4}\up{1.2} & \textbf{91.6}\up{0.8} & \textbf{93.4}\up{0.8} & \textbf{94.8}\up{0.8} & \textbf{95.8}\up{0.5} & \textbf{96.6}\up{0.4} & \textbf{89.5}\up{2.7} \\
\midrule
\multicolumn{11}{c}{\textbf{OlympiadBench \citep{he2024olympiadbenchchallengingbenchmarkpromoting}}} \\
\midrule
Base Model & 28.5 & 38.5 & 47.5 & 55.2 & 61.4 & 66.5 & 71.0 & 75.1 & 78.5 & 58.0 \\
REINFORCE & 38.2\up{9.7} & 43.8\up{5.3} & 49.0\up{1.5} & 53.7\dn{1.5} & 57.6\dn{3.8} & 60.9\dn{5.6} & 63.8\dn{7.2} & 66.3\dn{8.8} & 68.7\dn{9.8} & 55.8\dn{2.2} \\
GRPO \citep{shao2024deepseekmath} & \textbf{39.6}\up{11.1} & \textbf{45.9}\up{7.4} & 51.5\up{4.0} & 56.5\up{1.3} & 60.5\dn{0.9} & 63.8\dn{2.7} & 66.6\dn{4.4} & 69.1\dn{6.0} & 71.5\dn{7.0} & 58.3\up{0.3} \\
W-RF \citep{zhu2025the} & 38.2\up{9.7} & 45.0\up{6.5} & 51.3\up{3.8} & 57.0\up{1.8} & 61.7\up{0.3} & 65.6\dn{0.9} & 69.0\dn{2.0} & 72.0\dn{3.1} & 74.6\dn{3.9} & 59.4\up{1.4} \\
BBG (Ours) & 37.1\up{8.6} & 45.1\up{6.6} & \textbf{52.4}\up{4.9} & \textbf{58.8}\up{3.6} & \textbf{64.1}\up{2.7} & \textbf{68.3}\up{1.8} & \textbf{72.0}\up{1.0} & \textbf{75.7}\up{0.6} & \textbf{78.5}\zer & \textbf{61.3}\up{3.3} \\
\midrule
\multicolumn{11}{c}{\textbf{MinervaMath \citep{lewkowycz2022solvingquantitativereasoningproblems}}} \\
\midrule
Base Model & 24.4 & 34.8 & 45.3 & 54.5 & 61.9 & 67.9 & 72.6 & 76.6 & 80.9 & 57.6 \\
REINFORCE & 37.3\up{12.9} & 44.2\up{9.4} & 50.5\up{5.2} & 55.7\up{1.2} & 60.1\dn{1.8} & 63.9\dn{4.0} & 67.2\dn{5.4} & 70.0\dn{6.6} & 72.4\dn{8.5} & 57.9\up{0.3} \\
GRPO \citep{shao2024deepseekmath} & 39.2\up{14.8} & 45.2\up{10.4} & 50.4\up{5.1} & 55.1\up{0.6} & 59.9\dn{2.0} & 64.2\dn{3.7} & 68.0\dn{4.6} & 71.6\dn{5.0} & 74.6\dn{6.3} & 58.7\up{1.1} \\
W-RF \citep{zhu2025the} & \textbf{39.4}\up{15.0} & \textbf{46.2}\up{11.4} & \textbf{52.1}\up{6.8} & 57.6\up{3.1} & 62.7\up{0.8} & 67.0\dn{0.9} & 70.6\dn{2.0} & 74.0\dn{2.6} & 76.8\dn{4.1} & 60.7\up{3.1} \\
BBG (Ours) & 37.1\up{12.7} & 45.3\up{10.5} & 52.4\up{7.1} & \textbf{58.8}\up{4.3} & \textbf{64.5}\up{2.6} & \textbf{69.2}\up{1.3} & \textbf{73.3}\up{0.7} & \textbf{77.0}\up{0.4} & \textbf{81.1}\up{0.2} & \textbf{62.1}\up{4.5} \\
\midrule
\multicolumn{11}{c}{\textbf{AIME 2025 \citep{zhu2025the}}} \\
\midrule
Base Model & 7.4 & 11.4 & 16.0 & 20.9 & 26.4 & 31.8 & 36.9 & 43.2 & 53.3 & 27.5 \\
REINFORCE & 10.4\up{3.0} & 13.2\up{1.8} & 15.7\dn{0.3} & 18.5\dn{2.4} & 22.7\dn{3.7} & 28.7\dn{3.1} & 36.1\dn{0.8} & 43.9\up{0.7} & 50.0\dn{3.3} & 26.6\dn{0.9} \\
GRPO \citep{shao2024deepseekmath} & \textbf{10.8}\up{3.4} & \textbf{15.1}\up{3.7} & 19.1\up{3.1} & 22.6\up{1.7} & 26.3\dn{0.1} & 30.9\dn{0.9} & 36.2\dn{0.7} & 41.5\dn{1.7} & 46.7\dn{6.6} & 27.7\up{0.2} \\
W-RF \citep{zhu2025the} & 9.4\up{2.0} & 13.9\up{2.5} & 18.7\up{2.7} & 23.9\up{3.0} & 29.4\up{3.0} & 34.9\up{3.1} & 40.6\up{3.7} & 46.5\up{3.3} & 53.3\zer & 30.1\up{2.6} \\
BBG (Ours) & 9.9\up{2.5} & \textbf{15.1}\up{3.7} & \textbf{20.7}\up{4.7} & \textbf{26.2}\up{5.3} & \textbf{32.0}\up{5.6} & \textbf{38.5}\up{6.7} & \textbf{46.8}\up{9.9} & \textbf{56.3}\up{13.1} & \textbf{63.3}\up{10.0} & \textbf{34.3}\up{6.8} \\
\bottomrule
\end{tabular}
\vskip -0.2in
\end{table*}

\clearpage

\subsection{Further Analysis}
\label{sec:generalization}

\textbf{Answer extraction and verifier behavior.}
Our Pass@$k$ evaluation uses an \emph{any-boxed} extraction rule: a response is counted as correct if any boxed expression matches the ground truth after normalization. This relaxes only the answer-placement convention, not the correctness criterion. Empirically, this change improves the \emph{base model} across all benchmarks and also benefits trained models, especially on \emph{MinervaMath}, suggesting that some apparent errors are due to answer exposure and formatting rather than missing mathematical content. This also provides a useful lens on recent findings that RLVR can improve performance with extremely small or imperfect training signals \citep{wang2025reinforcement, shao2026spuriousrewardsrethinkingtraining}: a boxed-answer verifier rewards not only reasoning, but also producing answers in the form expected by the evaluator. Conversely, methods that prioritize output diversity may be disadvantaged by such format-sensitive protocols, because they optimize less directly for a single canonical answer presentation. We therefore interpret short-RLVR gains partly as improved instruction following and answer exposure, distinct from genuine reasoning-boundary expansion.

\textbf{Simple lessons from overtraining.}
BBG is derived from the Bayesian boundary utility (Eq.~\eqref{eq:boundary_utility}), which estimates how much an update on each problem can still contribute to the reasoning boundary, and weights updates accordingly. As shown in Table~\ref{table:math_aime_pass_k}, BBG attains the highest average Pass@$k$ on every benchmark, and is the only method whose Pass@$256$ matches or exceeds the base model across all four.

The same overtraining analysis also suggests two simpler approaches (Table~\ref{table:generalization}). The epoch-level overtraining identified in Section~\ref{sec:late_stage} motivates \emph{early stopping}, while the boundary saturation identified in Section~\ref{sec:immediate} motivates \emph{bucket masking}, which removes updates from high-success rollout buckets.
Early stopping at 5 epochs already recovers much of the high-$k$ Pass@$k$ lost in later training, e.g.\ raising MinervaMath Pass@$256$ from $77.9$ under full W-RF back to $79.4$, close to the base model's $80.9$, while retaining most of the Pass@$1$ gain. Bucket masking offers a complementary lever: masking $B_{\ge 6/8}$ preserves nearly all of the full-training Pass@$1$ while substantially improving Pass@$256$ on harder benchmarks relative to full W-RF. All three approaches stem from the same overtraining analysis and each balances Pass@$1$ and high-$k$ Pass@$k$ differently, allowing practitioners to choose according to their priorities.

\textbf{Efficiency.}
BBG spends backward computation only on problems with non-negligible estimated utility for the reference metric. Under the default setting, problems whose boundary utility falls below $\tau$ are gated out, and only active problems are accumulated into an effective batch. This reduces the amount of gradient computation spent on empirically saturated problems. At the same time, dropped problems still require rollouts in order to determine their bucket membership, and BBG may need to sample additional problems before enough active prompts are accumulated. Thus, BBG's efficiency benefit is primarily a reallocation of the expensive optimization step toward problems with non-negligible estimated contribution to the reference metric.


\begin{table*}[t]
\centering
\caption{Effect of early stopping and progressive bucket masking on the Pass@$1$\,/\,Pass@$256$\,/\,\emph{Avg.} trade-off. Format follows Table~\ref{table:math_aime_pass_k}.}
\vskip -0.07in
\label{table:generalization}
\small
\setlength{\tabcolsep}{1.0pt}
\renewcommand{\arraystretch}{1.0}
\setlength{\aboverulesep}{0.4ex}
\setlength{\belowrulesep}{0.4ex}
\resizebox{\textwidth}{!}{%
\begin{tabular}{l cccc}
\toprule
\textbf{Method} & \textbf{MATH-500} & \textbf{OlympiadBench} & \textbf{MinervaMath} & \textbf{AIME 2025} \\
\midrule
Base Model
  & 64.5\,/\,96.2\,/\,\colorbox{RefColor!10}{86.8}
  & 28.5\,/\,78.5\,/\,\colorbox{RefColor!10}{58.0}
  & 24.4\,/\,80.9\,/\,\colorbox{RefColor!10}{57.6}
  & 7.4\,/\,53.3\,/\,\colorbox{RefColor!10}{27.5} \\
\midrule
W-RF (20 ep)
  & \textbf{76.5}\up{12.0}\,/\,94.4\dn{1.8}\,/\,\colorbox{RefColor!10}{87.9\up{1.1}}
  & \textbf{38.5}\up{10.0}\,/\,73.7\dn{4.8}\,/\,\colorbox{RefColor!10}{59.4\up{1.4}}
  & \textbf{38.9}\up{14.5}\,/\,77.9\dn{3.0}\,/\,\colorbox{RefColor!10}{61.2\up{3.6}}
  & 9.7\up{2.3}\,/\,53.3\,/\,\colorbox{RefColor!10}{31.3\up{3.8}} \\
\;+ early stop (10 ep)
  & 75.6\up{11.1}\,/\,96.0\dn{0.2}\,/\,\colorbox{RefColor!10}{88.9\up{2.1}}
  & 37.3\up{8.8}\,/\,74.9\dn{3.6}\,/\,\colorbox{RefColor!10}{59.9\up{1.9}}
  & 37.6\up{13.2}\,/\,78.3\dn{2.6}\,/\,\colorbox{RefColor!10}{61.3\up{3.7}}
  & \textbf{10.2}\up{2.8}\,/\,56.7\up{3.4}\,/\,\colorbox{RefColor!10}{32.3\up{4.8}} \\
\;+ early stop (5 ep)
  & 74.0\up{9.5}\,/\,\textbf{96.6}\up{0.4}\,/\,\colorbox{RefColor!10}{88.9\up{2.1}}
  & 36.3\up{7.8}\,/\,77.4\dn{1.1}\,/\,\colorbox{RefColor!10}{60.7\up{2.7}}
  & 36.0\up{11.6}\,/\,79.4\dn{1.5}\,/\,\colorbox{RefColor!10}{61.6\up{4.0}}
  & 9.1\up{1.7}\,/\,53.3\,/\,\colorbox{RefColor!10}{29.6\up{2.1}} \\
\;+ mask $B_{8/8}$
  & 76.2\up{11.7}\,/\,94.4\dn{1.8}\,/\,\colorbox{RefColor!10}{88.4\up{1.6}}
  & 36.5\up{8.0}\,/\,77.6\dn{0.9}\,/\,\colorbox{RefColor!10}{60.5\up{2.5}}
  & 32.6\up{8.2}\,/\,80.2\dn{0.7}\,/\,\colorbox{RefColor!10}{61.4\up{3.8}}
  & 9.2\up{1.8}\,/\,60.0\up{6.7}\,/\,\colorbox{RefColor!10}{31.5\up{4.0}} \\
\;+ mask $B_{\ge 7/8}$
  & 74.5\up{10.0}\,/\,95.6\dn{0.6}\,/\,\colorbox{RefColor!10}{88.7\up{1.9}}
  & 37.6\up{9.1}\,/\,76.4\dn{2.1}\,/\,\colorbox{RefColor!10}{60.5\up{2.5}}
  & 34.8\up{10.4}\,/\,80.6\dn{0.3}\,/\,\colorbox{RefColor!10}{61.7\up{4.1}}
  & 9.8\up{2.4}\,/\,60.0\up{6.7}\,/\,\colorbox{RefColor!10}{31.3\up{3.8}} \\
\;+ mask $B_{\ge 6/8}$
  & 73.6\up{9.1}\,/\,96.0\dn{0.2}\,/\,\colorbox{RefColor!10}{88.8\up{2.0}}
  & 37.9\up{9.4}\,/\,77.2\dn{1.3}\,/\,\colorbox{RefColor!10}{60.7\up{2.7}}
  & 35.0\up{10.6}\,/\,80.8\dn{0.1}\,/\,\colorbox{RefColor!10}{61.7\up{4.1}}
  & 9.7\up{2.3}\,/\,53.3\,/\,\colorbox{RefColor!10}{31.1\up{3.6}} \\
\midrule
BBG (Ours)
  & 75.6\up{11.1}\,/\,\textbf{96.6}\up{0.4}\,/\,\colorbox{RefColor!10}{\textbf{89.5}\up{2.7}}
  & 37.1\up{8.6}\,/\,\textbf{78.5}\zer\,/\,\colorbox{RefColor!10}{\textbf{61.3}\up{3.3}}
  & 37.1\up{12.7}\,/\,\textbf{81.1}\up{0.2}\,/\,\colorbox{RefColor!10}{\textbf{62.1}\up{4.5}}
  & 9.9\up{2.5}\,/\,\textbf{63.3}\up{10.0}\,/\,\colorbox{RefColor!10}{\textbf{34.3}\up{6.8}} \\
\bottomrule
\end{tabular}%
}
\vskip -0.15in
\end{table*}

\section{Conclusion, Limitations, and Broader Impact}
We studied diversity collapse in RLVR through the lens of overtraining. We showed that under standard few-rollout setups, most updates target problems whose contribution to high-$k$ Pass@$k$ has already saturated, so the resulting policy sharpening degrades the reasoning boundary without expanding it. We further provided evidence that RLVR does acquire new reasoning capabilities during training, but this gain is masked by overtraining on already-solved problems. Building on this analysis, we proposed Bayesian Boundary Gating (BBG), which weights updates by their estimated marginal contribution to the reasoning boundary, and showed that it improves average Pass@$k$ across a wide range of $k$ on multiple benchmarks.
 
\textbf{Limitations.}
Our experiments are conducted on a single training set with 7B-scale models. Whether the same overtraining hold at larger model scales or on more diverse training mixtures remains to be verified. Diversity collapse likely has causes beyond the problem-level overtraining studied here, including token-level credit assignment difficulties, reward sparsity, and forms of diversity not captured by Pass@$k$. BBG's mechanism intentionally suppresses updates on high-success problems, which improves average Pass@k but comes at the cost of a notable Pass@1 reduction. In settings where single-sample accuracy is the primary metric, this trade-off may be undesirable.

\textbf{Broader Impact.}
This work aims to improve the reasoning diversity of LLMs. We do not foresee direct negative societal consequences; the methods studied here apply to mathematical reasoning benchmarks and do not introduce new risks beyond those inherent to LLMs themselves.

\clearpage
\appendix

\section{Training and Evaluation Details}
\label{app:training_details}

\subsection{Base Model and Training Data}
\label{app:models_data}

Our primary experiments use Qwen2.5-Math-7B \citep{yang2024qwen25mathtechnicalreportmathematical} as the base policy $\pi_{\theta_0}$. We train on the MATH training set \citep{hendrycks2021measuring}, which contains approximately 7,500 competition-level mathematics problems spanning seven subjects (Prealgebra, Algebra, Number Theory, Counting and Probability, Geometry, Intermediate Algebra, and Precalculus). Each problem is paired with a final answer in a canonical form that can be verified by exact string matching after normalization. Each training problem $x$ is formatted as a prompt following the instruction template of the base model. The verifier extracts the final boxed answer from the model's response and compares it against the ground truth using the normalization procedure from \citet{hendrycks2021measuring}.

\subsection{Training Infrastructure}
\label{app:training_infra}

All training runs use a single node with 4 NVIDIA H100 96GB GPUs. Evaluation runs use a single H100 GPU. We use the verl framework \citep{10.1145/3689031.3696075}, which decomposes the pipeline into rollout generation and policy optimization. On-policy rollouts are generated using vLLM \citep{10.1145/3600006.3613165} in bfloat16 precision with eager execution mode and GPU memory utilization set to $0.65$; tensor model parallelism is set to $1$. The actor model is trained with Fully Sharded Data Parallelism (FSDP) across all 4 GPUs with gradient checkpointing enabled.

\subsection{Shared Hyperparameters}
\label{app:hyperparams}

Table~\ref{tab:shared_hyperparams} lists the hyperparameters shared across all methods, following the practice of \citet{zhu2025the}. With approximately 7,500 training problems and a batch size of 1024, each epoch corresponds to roughly 7 training steps; the total schedule of 140 steps spans approximately 20 epochs. Following \citet{zhu2025the}, we set the KL penalty coefficient to $0$, as prior work has found KL regularization can be detrimental in the RLVR setting where the reward signal is exact.

\begin{table}[h]
\centering
\caption{Shared training hyperparameters across all methods.}
\label{tab:shared_hyperparams}
\small
\setlength{\tabcolsep}{8pt}
\begin{tabular}{ll}
\toprule
Hyperparameter & Value \\
\midrule
Rollouts per problem ($n$) & 8 \\
Training batch size (prompts) & 1024 \\
Mini-batch size (prompts) & 128 \\
PPO micro-batch size per GPU & 1 \\
Log-prob micro-batch size per GPU & 4 \\
Peak learning rate & $1 \times 10^{-6}$ \\
Learning rate schedule & Cosine decay \\
KL penalty coefficient & 0 \\
Maximum prompt length & 1024 tokens \\
Maximum response length & 3072 tokens \\
Rollout precision & bfloat16 \\
Total training steps & 140 ($\approx$ 20 epochs) \\
Checkpoint / validation frequency & Every 7 steps ($\approx$ 1 epoch) \\
\bottomrule
\end{tabular}
\end{table}

\subsection{Method-Specific Configurations}
\label{app:method_specific}

\paragraph{REINFORCE.}
Uses binary rewards $r(x,y) \in \{-1, +1\}$ with an exponential moving average baseline. No group-level normalization is applied; all $n=8$ rollouts contribute to the gradient regardless of within-group reward variance.

\paragraph{GRPO.}
GRPO \citep{shao2024deepseekmath} normalizes advantages within each problem's rollout group: $\hat{A}_i = (r_i - \bar{r})/(\sigma_r + \epsilon)$. 

\paragraph{W-RF.}
W-RF \citep{zhu2025the} scales the advantage for correct rollouts, while incorrect rollouts retain their full advantage weight. This asymmetric weighting partially mitigates diversity collapse by reducing the contribution of positive reinforcement.

\paragraph{BBG.}
BBG builds on W-RF ($w^+ = 0.3$) and adds problem-level gating based on the Bayesian boundary utility (Eq.~\eqref{eq:boundary_utility}). We set the boundary reference $k{=}16$, the geometric midpoint of the evaluated range $k \in [1,256]$, and the minimum weight ratio $\tau{=}0.01$. Table~\ref{tab:bbg_utility_per_bucket} lists the per-bucket utility under this configuration: buckets $B_{0/8}$, $B_{1/8}$, $B_{2/8}$, and $B_{3/8}$ receive nonzero weight ($\approx 67.1\%$, $23.3\%$, $7.4\%$, and $2.1\%$ respectively after active-weight normalization), while buckets $B_{\ge 4/8}$ are hard-gated and excluded from backpropagation. Because gating removes a substantial fraction of the training batch, BBG accumulates active problems into an effective batch of 512 prompts per gradient step.

\begin{table}[h]
\centering
\caption{Bayesian boundary utility $u(m;\,n{=}8,\,k{=}16)$ and normalized active weight for each rollout-success bucket under BBG.}
\label{tab:bbg_utility_per_bucket}
\small
\setlength{\tabcolsep}{5pt}
\begin{tabular}{cccc}
\toprule
Bucket $m/8$ & $u(m;\,8,16)$ & $\alpha(m) / \sum_j \alpha(j)$ & Gated? \\
\midrule
$B_{0/8}$ & 6.000 & 0.6711 & No \\
$B_{1/8}$ & 2.087 & 0.2334 & No \\
$B_{2/8}$ & 0.664 & 0.0743 & No \\
$B_{3/8}$ & 0.190 & 0.0212 & No \\
$B_{4/8}$ & 0.047 & 0 & Yes ($< \tau \cdot 9 = 0.09$) \\
$B_{5/8}$ & 0.010 & 0 & Yes \\
$B_{6/8}$ & 0.0017 & 0 & Yes \\
$B_{7/8}$ & $1.96 \times 10^{-4}$ & 0 & Yes \\
$B_{8/8}$ & $1.22 \times 10^{-5}$ & 0 & Yes \\
\bottomrule
\end{tabular}
\end{table}

\subsection{Evaluation Protocol}
\label{app:eval_protocol}

For each benchmark, we generate $N{=}256$ independent samples per problem with temperature $T{=}0.6$, nucleus sampling top-$p{=}0.95$, and a maximum generation length of 4096 tokens (longer than the training maximum of 3072 to avoid truncating reasoning chains). Inference uses vLLM in bfloat16 on a single GPU with batch size 1000.

Pass@$k$ is computed using the unbiased estimator of \citet{chen2021evaluatinglargelanguagemodels}: given $N$ total samples and $c$ correct samples for a problem,
\begin{equation}
\label{eq:passk_estimator}
\widehat{\text{Pass@}k} \;=\; 1 - \frac{\binom{N-c}{k}}{\binom{N}{k}}\,,
\end{equation}
which is unbiased for $1-(1-p)^k$ whenever $N \ge k$. We report Pass@$k$ for $k \in \{1, 2, 4, 8, 16, 32, 64, 128, 256\}$; the Avg.\ column is the arithmetic mean over these nine values. Answer extraction and normalization follow \citet{hendrycks2021measuring}.

\subsection{Evaluation Benchmarks}
\label{app:benchmark_details}

\emph{MATH-500} \citep{lightman2024lets} is a 500-problem subset of the MATH test set, serving as the in-distribution benchmark. The base model already achieves Pass@$256{=} 96.2\%$, leaving little room for observable boundary shrinkage. \emph{OlympiadBench} \citep{he2024olympiadbenchchallengingbenchmarkpromoting} contains 675 olympiad-level problems from international mathematics competitions. \emph{MinervaMath} \citep{lewkowycz2022solvingquantitativereasoningproblems} consists of 272 undergraduate-level quantitative reasoning problems spanning mathematics, physics, and related STEM disciplines. Both are out-of-distribution and contain many near-boundary problems, making them sensitive to diversity collapse. \emph{AIME 2025} \citep{zhu2025the} consists of the 30 problems from the 2025 American Invitational Mathematics Examination and is the most difficult benchmark; the overtraining regime for AIME likely begins at $k$ values beyond the evaluated range (consistent with \citealt{yue2025does}, who report degradation at $k{=}1024$).

\subsection{Further Experiment Details}
\label{app:ablation_details}

\paragraph{Bucket-restricted RLVR (Figure~\ref{fig:bucket_restricted}).}
We train variants restricting updates to a single bucket $B_{m/8}$ for $m \in \{0, 1, 4, 8\}$. Rollout outcomes are computed as usual, but only problems in the target bucket contribute to the gradient. All other hyperparameters are identical to W-RF.

\paragraph{Early stopping (Table~\ref{table:generalization}).}
We select the checkpoint at epoch 5 or 10 from the standard 20-epoch W-RF run; no additional tuning is performed.

\paragraph{Bucket masking (Table~\ref{table:generalization}).}
The masking variants (mask $B_{8/8}$, mask $B_{\ge 7/8}$, mask $B_{\ge 6/8}$) exclude problems in the specified high-success buckets from the gradient. Unlike BBG, these apply hard binary masks without utility-based weighting.

\paragraph{Boundary tracking (Figure~\ref{fig:boundary_tracking}).}
A problem is classified as solvable if it produces at least one correct response out of $N{=}256$ samples, and unsolvable otherwise. We identify problems with Pass@$256{=}0$ under the base policy and track their status at each checkpoint. Entry and exit counts in Figure~\ref{fig:boundary_tracking_entry_loss} are cumulative relative to the base classification; the sampling-noise analysis in Appendix~\ref{app:sampling_noise} addresses the concern that some transitions may be artifacts of finite-sample evaluation.

\subsection{Reproducibility}
\label{app:reproducibility}

All experiments use fixed random seeds for data shuffling. Rollout generation via vLLM is non-deterministic due to GPU parallelism, but the large number of rollouts and evaluation samples ensures that stochastic variation is small relative to the reported effect sizes. We report all numbers from a single training run per method.

\clearpage
\section{Analysis for the Zero-Success Bucket}
\label{app:zero_bucket}

This appendix expands the argument in Section~\ref{sec:immediate}: for high-\(k\) Pass@\(k\), the only rollout-success bucket that can still meaningfully contain boundary-expanding updates is the zero-success bucket \(B_{0/n}\). The statement should be understood at the resolution of the training signal: with only \(n\) rollouts per problem, using the empirical success rate as a plug-in estimate, even one correct rollout places the problem in an apparently saturated regime for Pass@\(k\).

For a problem with true single-sample success probability \(p\), its contribution to Pass@\(k\) is
\begin{equation}
J_k(p)=1-(1-p)^k.
\end{equation}
The remaining headroom is
\begin{equation}
1-J_k(p)=(1-p)^k,
\end{equation}
and the marginal gain from increasing \(p\) is
\begin{equation}
\frac{dJ_k(p)}{dp}=k(1-p)^{k-1}.
\end{equation}
Thus, \(J_k\) saturates exponentially fast in \(k\). In our main setting, \(k=256\) and \(n=8\). The smallest nonzero empirical success rate observed during training is \(\hat p=1/8\). If we plug this empirical rate into \(J_{256}\), we obtain
\begin{equation}
J_{256}(1/8)
=
1-\left(\frac{7}{8}\right)^{256}
=
1-1.43\times 10^{-15}
\approx 1.
\end{equation}
The corresponding marginal value is also essentially zero:
\begin{equation}
\left.\frac{dJ_{256}(p)}{dp}\right|_{p=1/8}
=
256\left(\frac{7}{8}\right)^{255}
\approx 4.18\times 10^{-13}.
\end{equation}
For \(m\ge 2\), the headroom is even smaller; for example,
\begin{equation}
1-J_{256}(2/8)=\left(\frac{3}{4}\right)^{256}\approx 1.04\times 10^{-32}.
\end{equation}
Therefore, once a problem has produced at least one correct rollout among the \(n=8\) training samples, its empirical contribution to Pass@\(256\) is already saturated. Further optimization on that problem may still increase Pass@\(1\), by making the sampled correct behavior more reliable, but it has essentially no room left to expand the high-\(k\) reasoning boundary.

This is why \(B_{0/n}\) is special. For \(m=0\), the empirical success rate is \(\hat p=0\), so the plug-in estimate gives
\begin{equation}
J_{256}(0)=0,
\qquad
\left.\frac{dJ_{256}(p)}{dp}\right|_{p=0}=256.
\end{equation}
At the level of the observed rollout group, \(B_{0/n}\) is the only bucket not already saturated for Pass@\(256\). If an update on such a problem moves even a small amount of probability mass toward an unobserved correct trajectory, the resulting improvement in \(J_{256}\) can be large.

However, \(B_{0/n}\) is also a noisy and tricky bucket. A problem can already be inside the Pass@\(256\) boundary while still producing zero successes in only \(n=8\) training rollouts. Indeed,
\begin{equation}
\Pr(\hat p_{\theta,n}=0\mid p)=(1-p)^n.
\end{equation}
Writing \(q=J_k(p)=1-(1-p)^k\), we can express this probability directly in terms of the Pass@\(k\) value:
\begin{equation}
\Pr(\hat p_{\theta,n}=0\mid J_k(p)=q)
=
(1-q)^{n/k}.
\end{equation}
For \(n=8\) and \(k=256\),
\begin{equation}
\Pr(\hat p_{\theta,8}=0\mid J_{256}(p)=q)
=
(1-q)^{1/32}.
\end{equation}
This probability remains large even when the problem is highly likely to be solved:
\begin{equation}
\begin{array}{c|c}
J_{256}(p) & \Pr(\hat p_{\theta,8}=0) \\ \hline
0.50 & 97.9\% \\
0.90 & 93.1\% \\
0.99 & 86.6\% \\
0.999 & 80.6\%
\end{array}
\end{equation}
Thus, \(B_{0/8}\) should not be interpreted as a clean set of truly unsolved problems. It should only be interpreted as the bucket that still contains substantial boundary-expanding opportunities under the coarse \(n=8\) training observation. It may also contain low-probability but already boundary-reachable problems. This explains both sides of the intervention in Section~\ref{sec:reframe}: restricting updates to \(B_{0/8}\) removes almost all empirically saturated updates, which helps Pass@\(256\), but the bucket itself is still noisy because \(n\ll k\).
One direct way to reduce this leakage is to increase the number of training rollouts \(n\). For a problem with true success probability \(p\), the probability that it leaks into the zero-success bucket is
\begin{equation}
\Pr(\hat p_{\theta,n}=0\mid p)=(1-p)^n .
\end{equation}
Thus, increasing \(n\) makes \(B_{0/n}\) a cleaner proxy for genuinely unresolved problems. For example, when \(p=0.05\), the leakage probability is \(0.95^8=66.3\%\) under \(n=8\), but drops to \(0.95^{16}=44.0\%\) under \(n=16\), and further to \(0.95^{32}=19.4\%\) under \(n=32\). Since \(p=0.05\) already corresponds to \(J_{256}(p)=1-0.95^{256}\approx 0.999998\), these are precisely the kinds of boundary-reachable problems that can contaminate the zero-success bucket under small rollout budgets.

\section{Sampling-Noise Analysis}
\label{app:sampling_noise}

This appendix quantifies how much of the boundary entry and loss observed in Section~\ref{sec:reframe} could be explained by finite-sample noise. Since Pass@\(k\) is estimated from a finite set of sampled completions \citep{chen2021evaluatinglargelanguagemodels}, the induced binary Pass@\(256\) boundary label is noisy at the problem level: a problem with a small but nonzero success probability may receive zero correct samples in one evaluation and at least one correct sample in another. We therefore distinguish weak one-success transitions from stronger multi-success transitions.

For problem \(i\) at checkpoint \(t\), let \(C_{i,t}\) be the number of correct responses among \(K=256\) independent evaluation samples. If the true single-sample success probability is \(p_{i,t}\), then
\begin{equation}
\Pr(C_{i,t}=c)
=
\binom{K}{c}p_{i,t}^{\,c}(1-p_{i,t})^{K-c},
\qquad c=0,1,\ldots,K.
\end{equation}
The empirical Pass@\(256\) boundary indicator is
\begin{equation}
Z_{i,t}=\mathbf{1}\{C_{i,t}>0\}.
\end{equation}
A boundary entry from the base model to checkpoint \(t\) is \(C_{i,0}=0\) and \(C_{i,t}>0\), while a boundary loss is \(C_{i,0}>0\) and \(C_{i,t}=0\).

\paragraph{Exact noise test for an observed entry.}
Suppose a problem has \(C_{i,0}=0\) under the base model and \(C_{i,t}=c\) at a later checkpoint. Under the null hypothesis that the true success probability did not change, \(H_0:p_{i,t}=p_{i,0}\), the \(c\) successes observed among the combined \(2K=512\) samples should be exchangeable between the two evaluations. Conditioning on the total number of successes, \(C_{i,0}+C_{i,t}=c\), removes the unknown success probability. Under \(H_0\), the \(c\) success positions are uniformly distributed among the \(2K\) sample positions. Therefore, the probability that all \(c\) successes fall in the later checkpoint is
\begin{equation}
P_{\mathrm{noise}}(c)
=
\Pr(C_{i,t}=c, C_{i,0}=0 \mid C_{i,0}+C_{i,t}=c, H_0)
=
\frac{\binom{K}{c}}{\binom{2K}{c}}.
\end{equation}
Equivalently, for \(K=256\),
\begin{equation}
P_{\mathrm{noise}}(c)
=
\frac{\binom{256}{c}}{\binom{512}{c}}
=
\prod_{j=0}^{c-1}\frac{256-j}{512-j}.
\end{equation}
This is the standard conditional form of Fisher's exact test for a \(2\times 2\) table \citep{dda9a3b2-7e66-38cd-8ce8-7d36cd60b878, agresti2013categorical}.
This gives the following exact one-sided sampling-noise scores:
\[
\begin{array}{c|cccccccc}
c & 1 & 2 & 3 & 4 & 5 & 6 & 7 & 10 \\ \hline
P_{\mathrm{noise}}(c)
& 0.5000 & 0.2495 & 0.1243 & 0.0618
& 0.0306 & 0.0152 & 0.0075 & 8.93\times 10^{-4}
\end{array}
\]
Thus, a one-success entry is weak evidence and can easily arise from sampling noise. By contrast, entries with \(c\ge5\) are unlikely under unchanged success probability at the \(5\%\) level, and entries with \(c\ge7\) are unlikely at the \(1\%\) level.
The same calculation applies to boundary loss. If \(C_{i,0}=c\) and \(C_{i,t}=0\), then under the unchanged-policy null,
\[
\Pr(C_{i,0}=c, C_{i,t}=0 \mid C_{i,0}+C_{i,t}=c, H_0)
=
\frac{\binom{K}{c}}{\binom{2K}{c}}.
\]
Therefore, losing a problem that had only one base correct sample is noisy, while losing a problem with many base correct samples is much less likely to be explained by resampling noise alone.

\paragraph{Observed entry under standard RLVR.}
We apply this calculation to the boundary-tracking experiment in Section~\ref{sec:reframe}. On the held-out benchmarks AIME 2025, MinervaMath, and OlympiadBench, there are \(15+57+193=265\) problems with \(C_{i,0}=0\) under the base model.

At epoch 2, \(70/265=26.4\%\) of these base-unsolved problems enter the empirical Pass@\(256\) boundary. Of these 70 entries, \(27\) have at least two correct samples and \(3\) have at least five correct samples; equivalently, these account for \(27/265=10.2\%\) and \(3/265=1.1\%\) of all base-unsolved problems. For the two intermediate-difficulty benchmarks emphasized in the main text, MinervaMath has \(18/57=31.6\%\) entries, of which \(9\) have \(c\ge2\) and \(2\) have \(c\ge5\). OlympiadBench has \(47/193=24.4\%\) entries, of which \(15\) have \(c\ge2\) and \(1\) has \(c\ge5\). Hence, many binary entries are indeed weak one-success events, but a non-negligible subset is supported by repeated correct samples.
At the final checkpoint, epoch 20 (step 140), \(38/265=14.3\%\) of the held-out base-unsolved problems remain inside the empirical boundary. Among these 38 remaining entries, \(17\) have \(c\ge2\), \(3\) have \(c\ge5\), and one has \(c\ge10\); as fractions of all base-unsolved problems, these are \(6.4\%\), \(1.1\%\), and \(0.4\%\), respectively.

Across all six evaluated checkpoints, \(134/265=50.6\%\) of the base-unsolved held-out problems enter at least once. Among them, \(64/265=24.2\%\) enter at least once with \(c\ge2\), \(11/265=4.2\%\) enter at least once with \(c\ge5\), and \(4/265=1.5\%\) enter at least once with \(c\ge7\). In addition, \(68/265=25.7\%\) enter in at least two consecutive evaluated checkpoints. These repeated and multi-success transitions are much harder to attribute to a single lucky sample.

\paragraph{Observed loss is also stronger than one-sample noise.}
The aggregate Pass@\(256\) decline is driven not only by noisy entry labels but also by boundary loss. At epoch 20, the held-out benchmarks have \(152\) losses among \(15+215+481=711\) base-solvable problems, i.e. \(152/711=21.4\%\). Many of these losses come from problems that had multiple correct samples under the base model: \(91\) had at least two base correct samples, and \(32\) had at least five base correct samples. Since \(P_{\mathrm{noise}}(5)=0.0306\), losses from problems with \(c\ge5\) base successes are unlikely to be explained by resampling noise alone at the \(5\%\) level.

Overall, at epoch 20 the held-out benchmarks have \(38\) entries but \(152\) losses, giving \(38-152=-114\) net boundary transitions. Over the \(976\) held-out problems, this corresponds to \(-114/976=-11.7\%\). This matches the observed aggregate Pass@\(256\) degradation: RLVR does move some initially unsolved problems into the boundary, but this gain is more than offset by losses on previously solvable problems.

\clearpage
\section{Additional Results for Section~\ref{sec:reframe}}
\label{app:boundary_dynamics}

This appendix provides additional results for Section~\ref{sec:reframe}.
In the main text, we show that standard RLVR can make initially unsolvable problems solvable, while aggregate Pass@$256$ may still decline because this gain is offset by boundary exit on previously solvable problems.
Here we provide per-benchmark Pass@$k$ curves, per-benchmark entry/loss statistics, and representative transition diagrams.

\begin{figure*}[h]
\centering
\includegraphics[width=\textwidth]{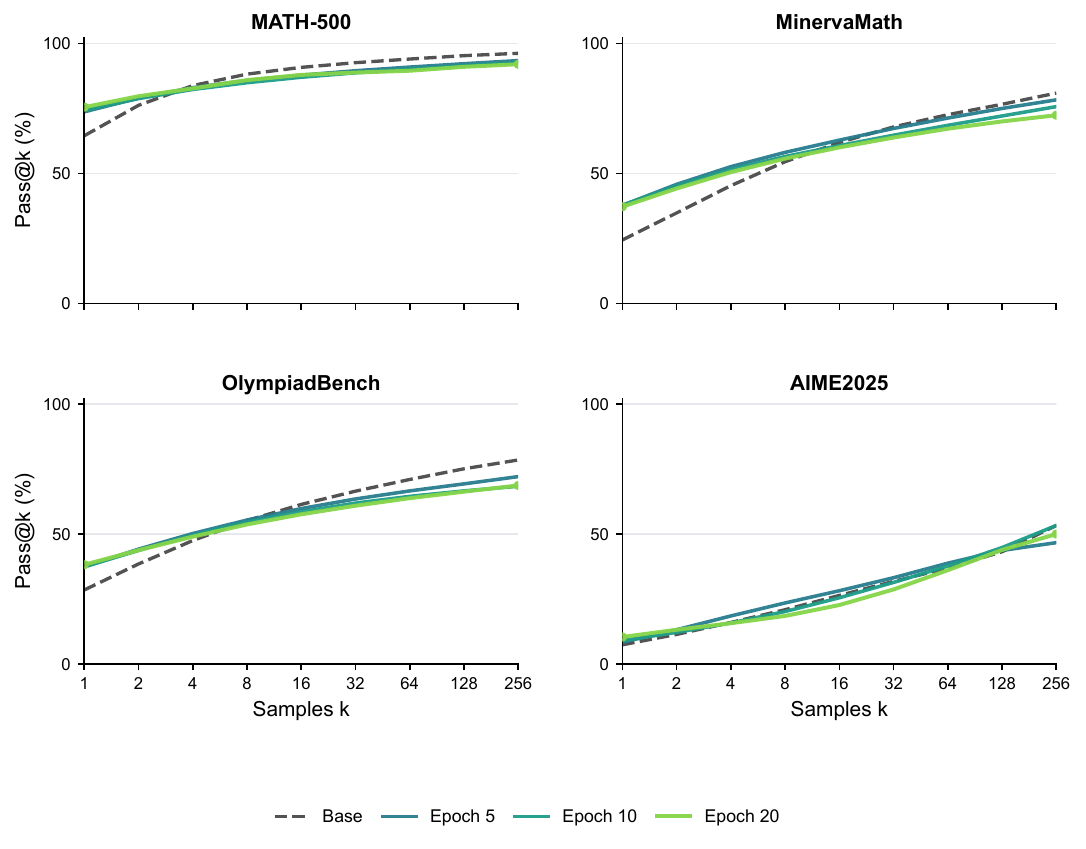}
\vskip -0.05in
\caption{
\emph{Per-benchmark Pass@$k$ curves across training.}
The averaged curve in Figure~\ref{fig:passk_curves_boundary_avg} is decomposed by benchmark.
On intermediate-difficulty benchmarks such as MinervaMath and OlympiadBench, later RLVR epochs improve or preserve low-$k$ performance while reducing high-$k$ Pass@$k$.
On easier or harder benchmarks, this effect is weaker because of ceiling effects or because the evaluated range of $k$ has not yet reached the saturated regime.
}
\label{fig:app_passk_curves_benchmarks}
\vskip -0.1in
\end{figure*}

\begin{figure*}[h]
\centering
\includegraphics[width=\textwidth]{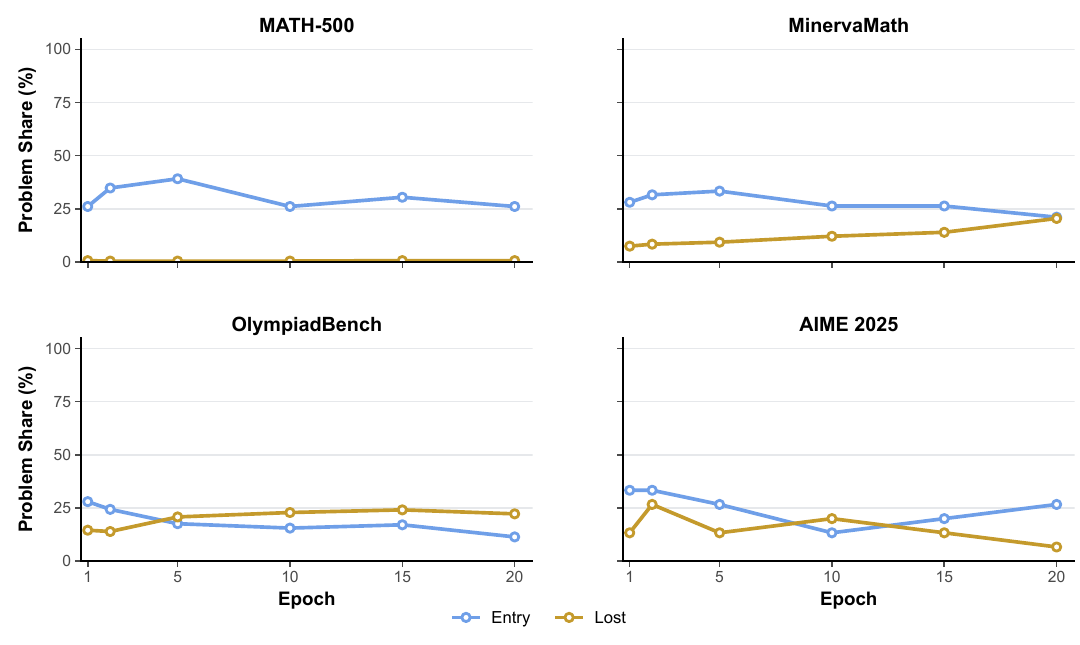}
\vskip -0.05in
\caption{
\emph{Per-benchmark boundary entry and exit under standard RLVR.}
Entry denotes the fraction of base-unsolved problems, i.e.\ problems with Pass@$256=0$ under the base policy, that become solvable within 256 samples at a later checkpoint.
Loss denotes the fraction of base-solvable problems that later become unsolved within 256 samples.
}
\label{fig:app_entry_lost_fraction_all_benchmarks}
\vskip -0.1in
\end{figure*}

\begin{figure*}[h]
\centering
\includegraphics[width=\textwidth]{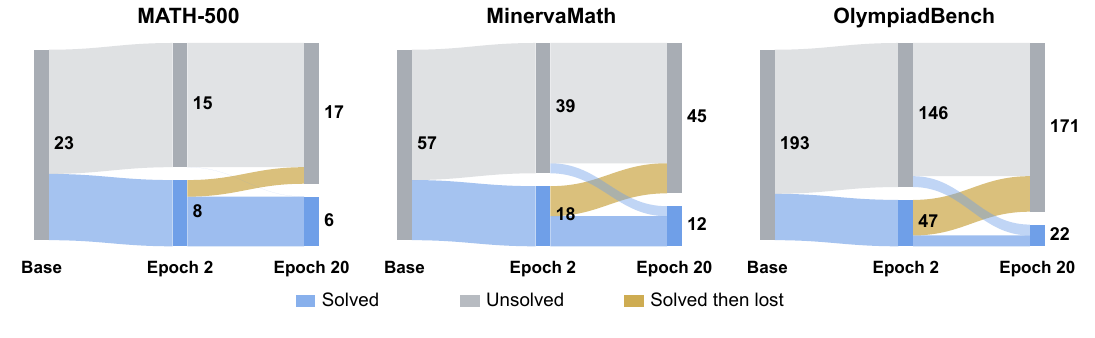}
\vskip -0.05in
\caption{
\emph{Representative boundary transitions for initially unsolved problems.}
For problems with Pass@$256=0$ under the base policy, we track whether they become solvable at epoch~2 and whether they remain solvable at epoch~20.
The diagram shows that standard RLVR can move initially unsolved problems into the boundary, but some of these newly reached problems are not retained through later training.
Thus, aggregate Pass@$256$ reflects the net effect of boundary entry and boundary exit, rather than the absence of newly acquired reasoning ability.
}
\label{fig:app_boundary_sankey_representative}
\vskip -0.1in
\end{figure*}

\end{document}